# Biomedical Reference Generation Remains Unreliable across 26 Large Language Models

Maxim Topaz[1,2,3*], Zhihong Zhang[1,2], Nir Roguin[4], Pallavi Gupta[1], Zichao Li[1], Laura-Maria Peltonen[5,6,7]

[1] *School of Nursing, Columbia University, New York, NY, USA*

[2] *Data Science Institute, Columbia University, New York, NY, USA*

[3] *VNS Health, New York, NY, USA*

[4] *Tel Aviv Sourasky Medical Center, Tel Aviv, Israel*

[5] *Department of Health and Social Management, University of Eastern Finland, Kuopio, Finland*

[6] *Wellbeing Services County of North Savo, Kuopio, Finland*

[7] *Wellbeing Services County of Southwest Finland, Turku, Finland*

[*] *Correspondence: Prof Maxim Topaz, School of Nursing, Columbia University, 560 West 168th Street, New York, NY 10032, USA. mt3315@cumc.columbia.edu*

## Abstract

**Background.** Large language models are increasingly used to help write biomedical text but may fabricate references to nonexistent work. How often large language models do so is not well characterized.

**Methods.** We prompted 26 language models from eight developers (2023 to 2026) to supply a missing reference for each of 69 biomedical passages across ten domains. References were classified as verifiable (real paper with a resolving identifier), partial matches (real paper without a resolving identifier), fabricated (no matching indexed paper), or declined (the model refused to supply a reference). A reference was considered correct in every evaluated bibliographic field only when it was verifiable and its journal, year, and listed authors matched those of the cited paper.

**Results.** Fabrication ranged from 10.2% (Claude Opus 4.8, which declined 52.1% of prompts) to 98.4% (Ministral 3B, which produced no verifiable reference). Claude Opus 4.6 and Claude Sonnet 4.5 produced similar proportions of verifiable references (77.6% and 76.6%) but named authors correctly in 78.7% and 28.7% of author-evaluable verifiable references, respectively, and were correct in every evaluated field in 54.6% and 19.9% of responses. GPT-5.5 was correct in every field in 48.1%. Across all models, 55.4% of responses were fabricated and 14.9% were correct in every field. Among the five tested models first released in 2026, the corresponding proportions were 35.3% and 31.8%, respectively.

**Conclusions.** Fabrication remained common, and no model was correct in every evaluated bibliographic field in more than 54.6% of responses. Models that identify real papers may still misstate their metadata, so references produced with model assistance require verification before use.

## Introduction

Language models are now embedded in biomedical writing: an analysis of full-text open-access papers estimated that 89% of biomedical papers published at the end of 2025 showed signs of language-model-assisted writing.[1] Writing assistance extends to reference lists, and references to papers that do not exist have been identified in the published biomedical literature.[2] Earlier evaluations of language models also found frequent fabricated references.[3–5] In our recent audit of the PubMed Central open-access subset, [2] we flagged thousands of suspected fabrications: references with plausible but nonexistent titles and author names paired with identifiers that resolve to unrelated papers; one paper on urinary-diversion surgery contained 18 fabricated references of 30. [2] That audit documented fabricated references at scale in the published record, but an audit of published papers cannot show which models produce them, at what rate, or in what form: whether a model invents a reference outright, declines to answer, or names a real paper and misstates its bibliographic details. Those distinctions determine where verification effort is needed at the moment of writing, before a fabricated reference enters the record. We therefore prompted 26 large language models to supply references for biomedical text passages and evaluated their bibliographic verifiability and metadata accuracy.

## Methods

We constructed 69 test passages from published biomedical papers across ten domains (cardiology, chemistry, diagnostics, endocrinology, immunology, infectious disease, neurology, oncology, pharmacology, and public health). In each passage, one reference was removed and replaced with a placeholder; 50 of the removed references were ones we had previously identified as fabricated in the published literature, and 19 were verified real references. Passages that had carried a fabricated reference in print allowed us to test whether the same textual context would elicit fabrication again; the 19 verified-reference passages served as controls. Each passage was presented to 26 language models from eight developers (OpenAI, Anthropic, Google, Meta, Mistral, DeepSeek, Alibaba, and xAI), accessed through application programming interfaces (APIs). The set spanned proprietary and open-weight systems, included several reasoning models, and had release dates from March 2023 to May 2026. Model inference was conducted in April and July 2026. Models received the full passage text containing the placeholder and were prompted to supply the complete missing reference, including title, authors, journal, year, and a PubMed identifier or digital object identifier (DOI). We tested the same two prompt variants in repeated runs and, where available, two temperature values; models without user-adjustable temperature were queried at the provider default. The individual model response was the unit of analysis. Sampling depth and passage coverage varied across models, so comparisons were descriptive (Table S1). As a sensitivity analysis, we standardized model-specific estimates to the intended mix of 50 passages with a fabricated reference and 19 with a verified reference. Reported estimates are observed values; standardized estimates appear only in sensitivity analyses. In total, model inference produced 24,518 classified responses; a further 359 empty or fragmentary responses were excluded as technical failures. Ninety-five percent confidence intervals for headline proportions were obtained by nonparametric bootstrap resampling of the 69 passages (10,000 replicates), which accounts for clustering of responses within passages; model comparisons remain descriptive.

Every generated title was searched against four bibliographic indexes: PubMed, Crossref, OpenAlex, and Google Scholar. After normalization of case and punctuation, title similarity was calculated with Python's difflib.SequenceMatcher; a ratio of 0.90 or higher defined a candidate match, and every candidate was manually reviewed to exclude a near-match to a different paper. The verification pipeline was deterministic; no language model was used to classify responses.

We classified each response into one of four categories. A response was verifiable when the generated title matched a real indexed paper and at least one supplied identifier, a PubMed identifier or a DOI, resolved to that same paper. A response was a partial match when the title matched a real indexed paper, but no supplied identifier resolved to it. A response was fabricated when the title matched no paper in the bibliographic search pipeline. A response was declined when the model refused to supply a reference. Verifiability is a weaker standard than bibliographic accuracy, because a reference can identify a real paper and still misstate it. We therefore compared the generated journal, year, and author list with the record of the cited paper, and counted a reference as correct in every evaluated field only when it was verifiable and all three matched. Correctness in every field is therefore an accuracy outcome within the verifiable category, not a fifth classification category. An author list was correct only when the generated names, in order, formed a prefix of the true list, each matching the true author at that position on surname and first initial after normalizing case, punctuation, and diacritics, so that a correct first author followed by invented coauthors was scored as an error; truncation with et al. was credited and group authorship was removed before matching. The 103 references that named no evaluable individual author were excluded from the author-only analysis, leaving 7,478; in the all-fields outcome, these 103 references were counted as incorrect. The study involved no human participants or identifiable private information and did not require institutional review board review.

## Results

Of the 24,518 classified responses, 13,580 (55.4%; 95% CI, 50.2 to 60.5) were fabricated and bore titles not found by the bibliographic search pipeline (Figure 1). Another 7,581 (30.9%) were verifiable and 2,948 (12.0%) were partial matches, and models declined in 409 responses (1.7%). Fabrication ranged from 10.2% (6.3 to 14.4) to 98.4% (96.4 to 99.7) across individual models. Among the five models first released in 2026, 35.3% (29.2 to 41.8) of responses were fabricated.

Among the 7,581 verifiable references, the journal was correct in 97.1% and the publication year in 95.8%, but the author list was correct in only 51.7% of the 7,478 author-evaluable references, with the remainder typically retaining a correct first author and inventing the remaining coauthors (Table 1). Taking all fields together, 48.3% of verifiable references were correct in every evaluated field. Among partial matches, the journal was correct in 67.7% and the year in 65.0%. Across all 24,518 responses, 14.9% (12.3 to 17.9) were correct in every field, and among the five models first released in 2026, 31.8% (26.0 to 37.6). The 48.3% figure is response-weighted; weighting every model equally lowered it to 45.2%, as several models with few verifiable references attributed authorship poorly.

**Table 1.** Accuracy of the 7,581 verifiable references by field.

A verifiable reference names a real indexed paper and supplies at least one identifier that resolves to it. Author accuracy is computed on the 7,478 verifiable references naming at least one evaluable individual author and requires that every listed name match the corresponding real author in order, with truncation by et al. credited. The 103 references without an evaluable individual author were counted as incorrect in the all-fields outcome. All fields together require the title, identifier, journal, year, and authors all to be correct. Across all 24,518 responses, 14.9% were correct in every field (30.9% verifiable × 48.3% correct in every field).

| Field | Correct (%) |
|---|---|
| Journal (n=7,581) | 97.1 |
| Publication year (n=7,581) | 95.8 |
| Author list (n=7,478) | 51.7 |
| All fields together (n=7,581) | 48.3 |

Verifiable references and bibliographically accurate references did not coincide at the model level. Claude Opus 4.6 (Anthropic) and Claude Sonnet 4.5 (Anthropic) produced almost identical shares of verifiable references, 77.6% and 76.6%, but Claude Opus 4.6 named the authors correctly in 78.7% (67.7 to 88.3) of them against 28.7% (19.2 to 38.8) for Claude Sonnet 4.5 and was correct in every field in 54.6% (41.8 to 66.9) of all its responses against 19.9% (12.9 to 27.5). Two further models showed the same dissociation: GPT-4o was verifiable in 47.1% of responses but correct on authors in 29.8%, and DeepSeek-V3.1 in 45.4% and 21.6%. GPT-5.5 (OpenAI) was verifiable in 75.8% of responses, correct on authors in 68.4%, and correct in every field in 48.1% (39.4 to 57.2). No model was correct in every field in more than 54.6% of its responses, and 11 of the 26 models were correct in fewer than one response in ten.

Some partial matches paired real titles with identifiers that resolved to unrelated papers. One model gave the exact title of a published cardiovascular adipose-tissue study, but a PubMed identifier for an orthodontics paper, and another produced a correct immunotherapy title with an identifier for an economics paper.

Verifiable references ranged from 77.6% to zero (Figure 1). In every developer that supplied models of different sizes, the smallest model tested produced the fewest verifiable references: Gemini 2.5 Flash reached 13.1% against 34.6% for Gemini 2.5 Pro (Google), Llama 4 Scout 2.2% against 36.5% for Llama 4 Maverick (Meta), and the three Qwen3 models (Alibaba) rose from 0.9% at 8 billion parameters to 12.9% at 235 billion. Ministral 3B (Mistral) fabricated 98.4% and was the only model producing no verifiable reference. Fabrication did not track whether a model was proprietary or open: DeepSeek-R1, the open-weight model with the highest verifiable share, fabricated 33.6%, less often than 10 of the 15 proprietary models, while OpenAI's own open-weight gpt-oss models fabricated 78.6% and 89.2%. No model eliminated fabrication.

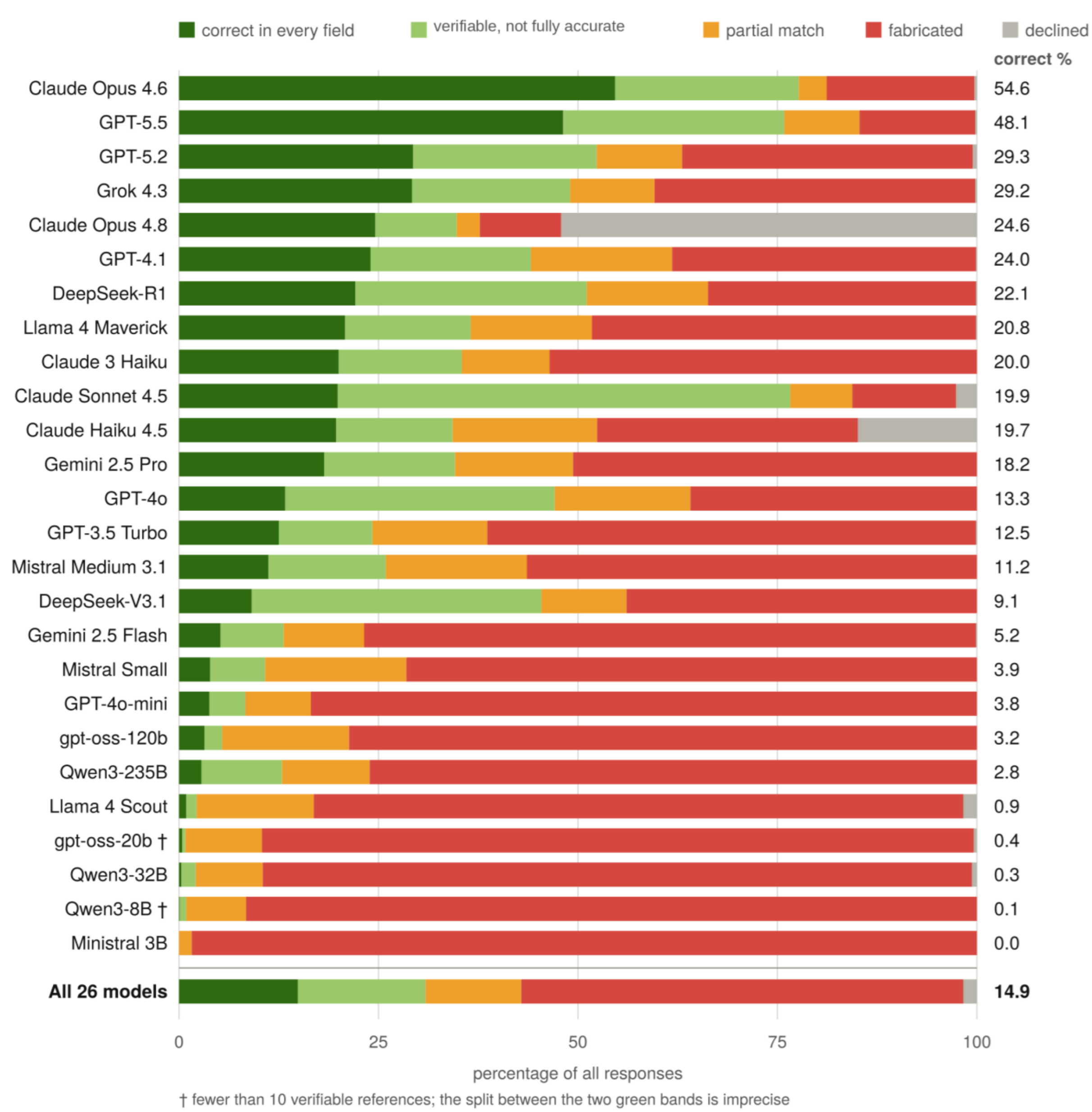

**Figure 1.** Bibliographic accuracy of references generated by 26 large language models.

Each bar shows the distribution of all responses received from a single model across up to 69 biomedical test passages and two prompt variants. Correct in every field: the title matched a real indexed paper, at least one supplied identifier resolved to that paper, and the journal, year, and listed author sequence met the prespecified matching rule. Verifiable: not fully accurate, the title and an identifier resolved to a real paper, but one or more of the remaining fields was wrong; in most cases, the first author was correct, and the remaining coauthors were invented. Partial match: the title matched a real indexed paper, but no supplied identifier resolved to it. Fabricated: the title was not found by the bibliographic search pipeline. Declined: the model refused to supply a reference. The two green bands together constitute the verifiable references. Bars sum to 100% and models are ordered by the proportion of responses correct in every field, shown at right.

Two models declined a substantial share of prompts. Claude Opus 4.8 (Anthropic) declined in 52.1% of responses, and Claude Haiku 4.5 (Anthropic) in 14.9%, both stating they could not supply reliable bibliographic information; every other model declined in fewer than 3% of responses, and most in none. Claude Opus 4.8 recorded the lowest fabrication rate on an all-response denominator, 10.2%, while declining half its prompts; among responses it answered, 21.3% were fabricated. Among models that declined in fewer than 3% of responses, Claude Sonnet 4.5 fabricated the least, at 13.0%.

For the 50 passages seeded with a known published fabrication, model outputs showed little resemblance to the original (mean title similarity 0.34 on a 0-to-1 scale; identifier overlap 0.1%). Models rarely reproduced components of the fabrications already circulating in the literature; when they fabricated a reference, they generally generated a different one.

Three sensitivity checks indicated that the main findings were not artifacts of analytic choices. Weighting each model equally (the unweighted mean of model-specific rates) rather than by its response count changed the fabrication rate only from 55.4% to 54.7%; passage-type standardization did not materially change model estimates or conclusions; and restricting verification to PubMed identifiers lowered the verifiable share from 30.9% to 18.0% but left fabrication unchanged at 55.4%.

**Discussion**

This study measured how often the tested large language models produced bibliographically accurate biomedical references. Most responses contained a fabricated reference naming no indexed paper, consistent with earlier reports of frequent fabrication. [3–5] Fabrication varied nearly tenfold across models, from 10.2% to 98.4%, so no single rate characterizes language models as a class; any aggregate reflects the models it pools. Among the five tested models first released in 2026, 35.3% of responses were fabricated and 31.8% were correct in every evaluated bibliographic field.

Verifiability is not bibliographic accuracy. A model that reliably located a real paper did not reliably describe it: Claude Sonnet 4.5 matched Claude Opus 4.6 on verifiable references but named the correct authors less than half as often, and GPT-4o and DeepSeek-V3.1 showed the same dissociation. Because the journal and year were correct in most verifiable references, authorship determined whether the bibliographic record was accurate, and its failure was systematic rather than random: models retained a correct first author and invented the remaining coauthors. Locating a source and attributing it are distinct capabilities, and a reference that names a real paper cannot be used without checking, an issue that pre-model quotation-accuracy audits did not confront.[6] These errors are often grouped together as hallucinations,[7,8] but outputs range from bibliographically accurate references, through real papers misattributed or lacking a resolving identifier, to plausible inventions assembled from real components.

Among all developers who supplied models of different sizes, the smallest model tested produced the fewest verifiable references, while open-weight versus proprietary status did not cleanly separate performance. These descriptive patterns are consistent with broader work showing that hallucination varies with model and generation conditions, but they do not isolate the effects of model size, training, or open-weight versus proprietary release.[8,9] The low overlap with published fabrications is consistent with de novo generation rather than reproduction but does not establish a shared mechanism.

One behavior stood apart from this gradient. Two models declined a substantial share of prompts, stating that they could not supply reliable bibliographic information, and abstention is appropriate when a citation cannot be grounded. Declining was inconsistent, however, appearing in only two of 26 models, and it changes how a fabrication rate should be read: Claude Opus 4.8 recorded the lowest all-response fabrication rate while answering about half the time. Methods that estimate a model's uncertainty in its own output offer one route to making such abstention systematic rather than incidental.[7]

This study has several limitations. First, our sample ranged from flagship systems to small open-weight models, so aggregate rates describe the tested set rather than a population of models in use. Second, sampling depth and passage coverage varied across models, so model comparisons are descriptive, although equal model weighting and passage-type standardization did not materially change the main findings. Decoding settings were also not identical: where temperature was not user-adjustable, the provider default applied, and defaults differ across providers, so results characterize models as deployed rather than under matched decoding conditions. Training-data recency and model capability are likewise confounded; newer models may perform better in part because their training data extends later. Third, models were tested through APIs without web search; results may differ for retrieval-enabled systems. Fourth, the prompts explicitly requested complete bibliographic fields, and rates under naturalistic use may differ. Fifth, fabricated in this study means unverifiable against four bibliographic indexes after manual review; a real but unindexed or heavily reworded title could in principle be misclassified, although the recovery of 548 real works outside the biomedical literature through OpenAlex indicates that the pipeline surfaces such cases. Finally, we evaluated bibliographic verifiability and metadata accuracy, not whether a real paper supported the associated passage. Because multiple papers may validly support the same claim, semantic relevance and degree of evidentiary support require expert review and should be examined in future work.

These findings indicate how the problem might be reduced. Reference errors are common in the published, peer-reviewed literature,[6] so references produced with language-model assistance warrant verification before use. Coupling reference generation with real-time database retrieval may reduce fabrication by grounding the output in an existing record.[10] Models already name a real indexed paper in more than four of every ten responses, so reliable retrieval could convert many into valid references, and the failure of author attribution suggests retrieval should supply the full record rather than leave the model to complete it.

## Code and data availability

The prompt texts, the model version list, the scoring script, and aggregated per-model results will be deposited in a public repository (Zenodo); the deposit identifier will be added at proof. Derived data underlying the reported results are available from the corresponding author on reasonable request. Source passages and complete model outputs are not publicly available because their release could identify published papers containing fabricated references.

## Funding

This work received no specific funding.

## Author contributions

M.T. conceived the study, designed the experimental protocol, developed the automated verification pipeline, conducted all API experiments, analyzed the data, and wrote the manuscript. N.R., P.G., Z.Z., Z.L., and L.-M.P. contributed to the study design, participated in data analysis, and critically revised the manuscript. All authors reviewed and approved the final manuscript.

## Competing interests

M.T. reports that this study used commercial API services from multiple model developers, with open-weight models accessed through a unified inference API, for data generation. No author has a financial relationship with, holds equity in, or received funding or in-kind support from any of the model developers evaluated in this study. The authors declare no other competing interests.

## Use of AI

During preparation of this work, the authors used Claude (Anthropic) to assist with editing the manuscript and with writing analysis code. All study design decisions, response classifications, and interpretations were made and supervised by the authors, and the reference classification pipeline was deterministic and did not use a language model. The authors reviewed and verified all code, analyses, and text and take full responsibility for the content of the publication.

# Appendix

**Table S1.** Reference verification results for each of the 26 large language models, computed across 24,518 classified responses; 359 empty or fragmentary responses were excluded as technical failures. Verifiable, partial match, fabricated, and declined are each computed on all responses received from that model and sum to 100%. Per-model response counts ranged from 382 to 1,378; three models completed a partial set of the 69 passages (48 to 53), so denominators vary by model. Response counts differ because run counts, prompt variants, and temperature conditions varied across models. Released: month and year of public API availability. Verifiable: the generated title matches a real indexed paper and at least one supplied identifier resolves to that paper. Partial match: the title matches a real indexed paper, but no supplied identifier resolves to it. Fabricated: the title was not found by the bibliographic search pipeline. Declined: the model refused to supply a reference. Responses (n) is the number of classified responses used as the denominator for the four outcome percentages. Authors correct is the percentage of the model's author-evaluable verifiable references whose listed author sequence met the prespecified matching rule and is not reported for models with fewer than 10 verifiable references. Models are ordered by release date.

| Model | Developer | Released | Verifiable (%) | Partial match (%) | Fabricated (%) | Declined (%) | Responses (n) | Authors correct (%) |
|---|---|---|---|---|---|---|---|---|
| GPT-3.5 Turbo | OpenAI | Mar 2023 | 24.2 | 14.4 | 61.2 | 0.1 | 1,378 | 53.9 |
| Claude 3 Haiku | Anthropic | Mar 2024 | 35.4 | 11.0 | 53.5 | 0.0 | 1,377 | 57.9 |
| GPT-4o | OpenAI | May 2024 | 47.1 | 17.0 | 35.9 | 0.0 | 1,360 | 29.8 |
| GPT-4o-mini | OpenAI | Jul 2024 | 8.3 | 8.3 | 83.4 | 0.0 | 1,333 | 46.8 |
| DeepSeek-R1 | DeepSeek | Jan 2025 | 51.1 | 15.2 | 33.6 | 0.1 | 673 | 46.0 |
| Gemini 2.5 Pro | Google | Mar 2025 | 34.6 | 14.8 | 50.6 | 0.0 | 1,374 | 55.6 |
| GPT-4.1 | OpenAI | Apr 2025 | 44.1 | 17.7 | 38.1 | 0.1 | 1,333 | 55.8 |
| Llama 4 Maverick | Meta | Apr 2025 | 36.5 | 15.2 | 48.1 | 0.1 | 788 | 60.6 |
| Llama 4 Scout | Meta | Apr 2025 | 2.2 | 14.7 | 81.4 | 1.7 | 764 | 43.8 |
| Qwen3-32B | Alibaba | Apr 2025 | 2.1 | 8.4 | 89.0 | 0.6 | 725 | 20.0 |
| Qwen3-8B | Alibaba | Apr 2025 | 0.9 | 7.5 | 91.7 | 0.0 | 805 | not reported |
| Gemini 2.5 Flash | Google | Apr 2025 | 13.1 | 10.1 | 76.7 | 0.1 | 1,241 | 43.0 |
| Qwen3-235B | Alibaba | Jul 2025 | 12.9 | 11.0 | 76.1 | 0.0 | 819 | 24.0 |
| DeepSeek-V3.1 | DeepSeek | Aug 2025 | 45.4 | 10.7 | 43.9 | 0.0 | 822 | 21.6 |
| Mistral Medium 3.1 | Mistral | Aug 2025 | 25.9 | 17.7 | 56.4 | 0.0 | 814 | 45.9 |
| gpt-oss-120b | OpenAI | Aug 2025 | 5.4 | 15.9 | 78.6 | 0.0 | 810 | 61.9 |
| gpt-oss-20b | OpenAI | Aug 2025 | 0.8 | 9.6 | 89.2 | 0.4 | 788 | not reported |
| Claude Sonnet 4.5 | Anthropic | Sep 2025 | 76.6 | 7.8 | 13.0 | 2.6 | 991 | 28.7 |
| Claude Haiku 4.5 | Anthropic | Oct 2025 | 34.3 | 18.2 | 32.7 | 14.9 | 986 | 61.8 |
| GPT-5.2 | OpenAI | Dec 2025 | 52.3 | 10.7 | 36.4 | 0.5 | 1,173 | 59.7 |
| Ministral 3B | Mistral | Dec 2025 | 0.0 | 1.6 | 98.4 | 0.0 | 799 | not applicable |
| Claude Opus 4.6 | Anthropic | Feb 2026 | 77.6 | 3.5 | 18.5 | 0.3 | 933 | 78.7 |
| Mistral Small | Mistral | Mar 2026 | 10.8 | 17.7 | 71.4 | 0.0 | 823 | 42.4 |
| GPT-5.5 | OpenAI | Apr 2026 | 75.8 | 9.4 | 14.5 | 0.2 | 414 | 68.4 |
| Grok 4.3 | xAI | Apr 2026 | 49.0 | 10.6 | 40.2 | 0.2 | 813 | 63.5 |
| Claude Opus 4.8 | Anthropic | May 2026 | 34.8 | 2.9 | 10.2 | 52.1 | 382 | 73.5 |
| **All 26 models** | | | **30.9** | **12.0** | **55.4** | **1.7** | **24,518** | **51.7** |